\documentclass{midl} % Include author names

\usepackage{url}
\usepackage{enumitem}
\setlist{nosep, leftmargin=14pt}

\usepackage{mwe} % to get dummy images
\usepackage{indentfirst}          
\usepackage{placeins}       
\usepackage{booktabs}
\usepackage{multirow}
\usepackage{caption}
\usepackage{float}      % for [H]
\usepackage{placeins}   % for \FloatBarrier

\usepackage{booktabs,caption}

\usepackage[letterpaper,
  left=1.03in, right=1.03in,   % +0.03in để tổng width <= 6.44in (đảm bảo ≥ 1.00in thực đo)
  top=0.80in, bottom=1.00in,
  includeheadfoot, heightrounded
]{geometry}

\usepackage{mwe} % to get dummy images
\jmlrvolume{-- Under Review}
\jmlryear{2026}
\jmlrworkshop{Full Paper -- MIDL 2026 submission}
\editors{Under Review for MIDL 2026}

 \title[Short Title]{DualProtoSeg: Simple and Efficient Design with Text- and Image-Guided Prototype Learning for Weakly Supervised Histopathology Image Segmentation}

\midlauthor{
\Name{Anh M. Vu\midljointauthortext{Contributed equally}\nametag{$^{1}$}} 
  \Email{mvu9@cougarnet.uh.edu} \AND
  \Name{Khang P. Le\midlotherjointauthor\nametag{$^{2}$}}
  \Email{khang.lekhangle23@hmcut.edu.vn} \AND
  \Name{Trang T. K. Vo\midlotherjointauthor\nametag{$^{3}$}}
  \Email{trangvtk.18@grad.uit.edu.vn} \AND 
  \Name{Ha Thach \nametag{$^{4}$}}
  \Email{nguyenbichha.thach@student.uts.edu.au} \AND
  \Name{Huy Hung Nguyen\nametag{$^{5}$}}
  \Email{25hung.nh@vinuni.edu.vn} \AND
  \Name{David Yang\nametag{$^{6}$}}
  \Email{dongjun.yang@emory.edu} \AND
  \Name{Han H. Huynh\nametag{$^{7}$}}
  \Email{m658112001@tmu.edu.tw} \AND
  \Name{Quynh Nguyen\nametag{$^{1}$}}
  \Email{qtnguy50@CougarNet.UH.EDU} \AND
  \Name{Tuan M. Pham\nametag{$^{1}$}}
  \Email{tmpham42@cougarnet.uh.edu} \AND
  \Name{Tuan-Anh Le\nametag{$^{1}$}}
  \Email{tle99@CougarNet.UH.EDU} \AND
  \Name{Minh H. N. Le\nametag{$^{8}$}}
  \Email{Johnminhle@ieee.org} \AND
  \Name{Thanh-Huy Nguyen\nametag{$^{9}$}}
  \Email{thanhhun@andrew.cmu.edu} \AND
  \Name{Akash Awasthi\nametag{$^{1}$}}
  \Email{aawasth3@CougarNet.UH.EDU} \AND
  \Name{Chandra Mohan\nametag{$^{1}$}}
  \Email{cmohan@Central.UH.EDU} \AND
  \Name{Zhu Han\nametag{$^{1}$}}
  \Email{zhan2@Central.UH.EDU} \AND
  \Name{Hien Van Nguyen\nametag{$^{1}$}}
  \Email{hvnguy35@Central.UH.EDU} \\[2ex]
  \addr $^{1}$ University of Houston, Houston, TX, USA \\
  \addr $^{2}$ Ho Chi Minh City University of Technology, Vietnam \\
  \addr $^{3}$ University of Information Technology, Ho Chi Minh City, Vietnam \\
  \addr $^{4}$ University of Technology Sydney, Australia \\
  \addr $^{5}$ Vin University, Hanoi, Vietnam \\
  \addr $^{6}$ Department of Computer Science, Emory University, Atlanta, GA, USA \\
  \addr $^{7}$ College of Medical Science and Technology, Taipei Medical University, Taipei, Taiwan \\
  \addr $^{8}$ Montefiore Medical Center, Albert Einstein College of Medicine, Bronx, NY, USA \\
  \addr $^{9}$ School of Computer Science, Carnegie Mellon University, Pittsburgh, PA, USA \\  % ← THIS LINE WAS MISSING
}

\begin{document}
\maketitle

\begin{abstract}
Weakly supervised semantic segmentation (WSSS) in histopathology seeks to reduce annotation cost by learning from image-level labels, yet it remains limited by inter-class homogeneity, intra-class heterogeneity, and the region-shrinkage effect of CAM-based supervision. We propose a simple and effective prototype-driven framework that leverages vision–language alignment to improve region discovery under weak supervision. Our method integrates CoOp-style learnable prompt tuning to generate text-based prototypes and combines them with learnable image prototypes, forming a dual-modal prototype bank that captures both semantic and appearance cues. To address oversmoothing in ViT representations, we incorporate a multi-scale pyramid module that enhances spatial precision and improves localization quality. Experiments on the BCSS-WSSS benchmark show that our approach surpasses existing state-of-the-art methods, and detailed analyses demonstrate the benefits of text description diversity, context length, and the complementary behavior of text and image prototypes. These results highlight the effectiveness of jointly leveraging textual semantics and visual prototype learning for WSSS in digital pathology. Code is available at: \url{https://github.com/maianhpuco/DualProtoSeg.git}

% Weakly-supervised semantic segmentation (WSSS) remains challenging when only image-level labels are available. We propose a one-stage, cluster-free framework that yields high-quality pseudo masks via a text-guided, multi-prototype, hierarchical design: input images go through a SegFormer backbone for multi-scale features, while class descriptions are encoded by CLIP to semantic text embeddings. For each class, learnable visual prototypes are adaptively projected to all scales, and cosine similarity between prototypes and image features produces hierarchical CAMs, which are fused CAMs (with fixed weights) and optionally refined with DenseCRF to obtain pseudo masks. During training, a contrastive alignment loss pulls prototypes toward the text embeddings of active classes, and a Prototype Diversity Regularizer using Jeffrey’s divergence plus a context-guided attention weighting mechanism prevents prototype collapse and boosts localization accuracy. The resulting pipeline produces accurate pseudo masks in a single forward pass and outperforms previous prototype-based methods on BCSS-WSSS dataset while preserving simplicity and efficiency.

\end{abstract}


\section{Introduction}
% \textcolor{red}{1. WSSSS: why we need prototype based segmentation in histopathology } \\
% \textcolor{red}{2. WSSS using CLIP based? why CLIP-based? current pipeline and draw back? FrozenClip (2 version)} \\
% \textcolor{red}{3. CLIP-based foundation model in Hispathology}  (text - image alignment) \\ 
% \textcolor{red}{4. Current pipeline for WSSS using CLIP-based foundation model and draw back (optional)}  \\
% \textbf{blue}{Contribution} \\
% \textbf{2 pages}

Weakly supervised segmentation (WSS) has become a practical alternative to costly pixel-level annotation in histopathology~\cite{Xiangbin_Liu,Minaee}, relying instead on weak labels such as image-level tags~\cite{Zhi-Hua_Zhou}. Most WSS follows a two-stage paradigm: a classifier trained on weak labels generates pseudo-masks (e.g., CAMs or instance labels), which then supervise a fully supervised segmentation model. CAMEL~\cite{CAMEL}, for instance, converts MIL-derived instance labels into approximate pixel masks for downstream segmentation. Image-level WSSS methods such as TransWS~\cite{Shaoteng_Zhang_TransWS} use a classification backbone to produce coarse localization cues and refine them through a transformer-based decoder, while TPRO~\cite{Shaoteng_Zhang_TPRO} leverages class descriptions as text prompts and a knowledge-attention module to build pseudo-masks. Other weak annotation strategies include sparse point methods~\cite{Qu_Spatial}, which expand point labels into coarse masks and refine boundaries via CRF \cite{CRF}, and bounding-box–based methods~\cite{Bounding_boxes}, which enforce tightness and background constraints directly in the loss function, enabling training with only box labels.

Class-activation-based (Grad-CAM-style) WSSS remains one of the most widely used paradigms: a classification network is trained with image-level labels, and its gradient-based activation maps are used as pseudo-masks for segmentation. However, CAMs inherently highlight only the most discriminative regions that support the image-level prediction, rather than representing the full spatial extent of each tissue type. This limitation is particularly problematic in histopathology, where intra-class heterogeneity (large morphological variation within a class) and inter-class homogeneity (high visual similarity between different classes) are pervasive. As a result, CAMs tend to collapse onto a small subset of visually distinctive patterns and systematically miss large portions of relevant structures.  To compensate for these weaknesses, prior works have proposed multi-scale CAM fusion~\cite{Multi-Scale_CAM}, instance- or region-aware activation mechanisms~\cite{Instance-level_CAM}, and additional priors such as self-supervision, background suppression, saliency, and vision–language guidance~\cite{CAM_Prior}. While these strategies improve coverage and boundary quality, they still operate within the inherent constraints of activation-based localization, and thus remain limited in their ability to model the full morphological diversity required for reliable histopathology WSSS. Recent attribution methods such as CIG ~\cite{vu2025cig} show that stronger class-discriminative cues exist beyond CAMs, but they remain interpretability tools rather than sources of dense masks, reinforcing the need for richer representation mechanisms.

Prototype-based WSSS offers a promising remedy to the limitations of CAM-based methods because prototypes capture characteristic morphological patterns rather than relying solely on highly discriminative regions. By modeling multiple representative appearances per class, prototype systems naturally mitigate intra-class heterogeneity and reduce confusion arising from inter-class homogeneity. Clustering-based variants such as PBIP~\cite{PBIP} and SIPE~\cite{SIPE} estimate semantic centers using K-means or affinity propagation and improve region coverage through contrastive prototype matching. Learnable prototype approaches, including ProtoPNet~\cite{ProtoPNet}, ViLa-MIL~\cite{ViLa_MIL}, PIP-Net~\cite{PIP_Net}, and LDP~\cite{LDP}, extend this idea by learning prototypes directly from data. Among them, LDP is a fully learnable prototype–based WSSS framework related to ours but fundamentally orthogonal, as it focuses on prototype refinement within a visual-only setting, whereas our work leverages text-image alignment to introduce multimodal semantic cues. Recent developments further enhance prototype robustness:~\cite{Prototype_with_CLIP} aligns visual prototypes with text-derived semantics, ProtoSeg~\cite{ProtoSeg} introduces a diversity constraint to prevent prototype collapse, and HisynSeg~\cite{HisynSeg} enforces representational diversity via synthetic mixing to improve pseudo-mask completeness. 

Despite these advances, prototype-based WSSS still faces key limitations. (1) Clustering-based methods introduce substantial computational overhead and are sensitive to hyperparameters. (2) Learnable prototype frameworks rarely leverage pathology-specific vision–language pretraining, missing an opportunity to incorporate rich semantic cues from paired image–text datasets. (3) Most prior work relies solely on visual prototypes, lacking the complementary semantic grounding that text–image alignment can provide. These gaps highlight the need for a more efficient, multimodal prototype framework that is robust to morphological variability and capable of leveraging modern pathology foundation models. 

 Modern histopathology encoders generally fall into two categories: vision–language models trained on large-scale pathology image–text pairs, and vision-only ViTs trained with powerful self-supervised objectives. Multimodal models such as CONCH~\cite{CONCH} and QuiltNet capture pathology-specific image–text alignment and demonstrate strong zero-shot generalization, while vision-only models like UNI~\cite{UNI}, CTransPath~\cite{CTransPath}, and GigaPath~\cite{GigaPath} provide robust visual representations learned from extensive SSL pretraining. However, most downstream WSSS pipelines treat these encoders purely as visual feature extractors and overlook both their multimodal capabilities and their hierarchical intermediate representations.

Inspired by advances in prototype learning and the strong capabilities of pathology-focused vision–language models for weak supervision, we propose \textbf{DualProtoSeg}, a simple and efficient framework that unifies text-guided and image-guided prototype learning for clustering-free WSSS in histopathology. Whole-slide patches are processed by a CLIP-style backbone (e.g., CONCH) to extract multi-scale visual features, while class descriptions are encoded by the text encoder and refined through learnable prompt tuning. The resulting image and text prototypes are projected across feature scales, and cosine similarity with spatial features produces hierarchical CAMs that are fused into dense pseudo-masks. A semantic alignment loss pulls visual prototypes toward the corresponding text embeddings, and a diversity regularizer prevents prototype collapse, improving spatial coverage and localization. This complementary multimodal design mitigates oversmoothing in ViT-based backbones and produces accurate segmentation masks under weak supervision.

Our main contributions are as follows:
\begin{enumerate}
    \item We introduce a weakly supervised segmentation framework that leverages pretrained CLIP-based vision-language representations, using text–image alignment as an additional source of semantic guidance.
    \item We propose a dual-prototype architecture that jointly models visual and textual concepts, enabling complementary appearance- and semantics-driven pseudo-mask generation.
    \item We incorporate a learnable prompt module in the text branch, allowing class descriptions to adapt to the dataset and providing stronger multimodal supervision.
\end{enumerate}

  \

\section{Method}

\textbf{Overview.} Dual-ProtoSeg performs weakly supervised semantic segmentation by combining multi-scale visual features with dual-modal prototypes. The frozen CONCH ViT-B/16 ~\cite{CONCH} encoder provides intermediate image features that are refined into a multi-scale pyramid, while the text branch encodes class descriptions using learnable context tokens. Text and image prototypes are projected to each scale and compared with spatial features via cosine similarity to generate class activation maps (CAMs). The multi-scale CAMs are upsampled, fused into pseudo-masks, and supervised with image-level classification losses. A DenseCRF ~\cite{CRF} further refines the fused CAM at inference to produce the segmentation mask.
% \textbf{Overview.} DualProtoSeg consists of three main components: (1) a multi-scale image branch, (2) a text branch with learnable prompt tuning, and (3) a dual-modal prototype module used for similarity mapping and prediction. As illustrated in Figure~\ref{fig:overview}, the image branch extracts intermediate hidden states from a pathology-specific vision–language encoder (e.g., CONCH) and refines them into a multi-scale feature pyramid to improve spatial localization. In parallel, the text branch encodes class descriptions using the frozen text encoder and adapts them with learnable context tokens, producing text-guided prototypes that capture semantic class cues. These text prototypes are combined with learnable image prototypes to form a unified dual-modal prototype bank, which is projected to each feature scale. Cosine similarity between prototypes and spatial features generates hierarchical CAMs used for both segmentation and classification. The multi-scale CAMs are fused into pseudo-masks and refined during inference with DenseCRF. The following subsections detail each component of the framework.

\begin{figure}[h]
    \centering
    \includegraphics[width=1\linewidth]{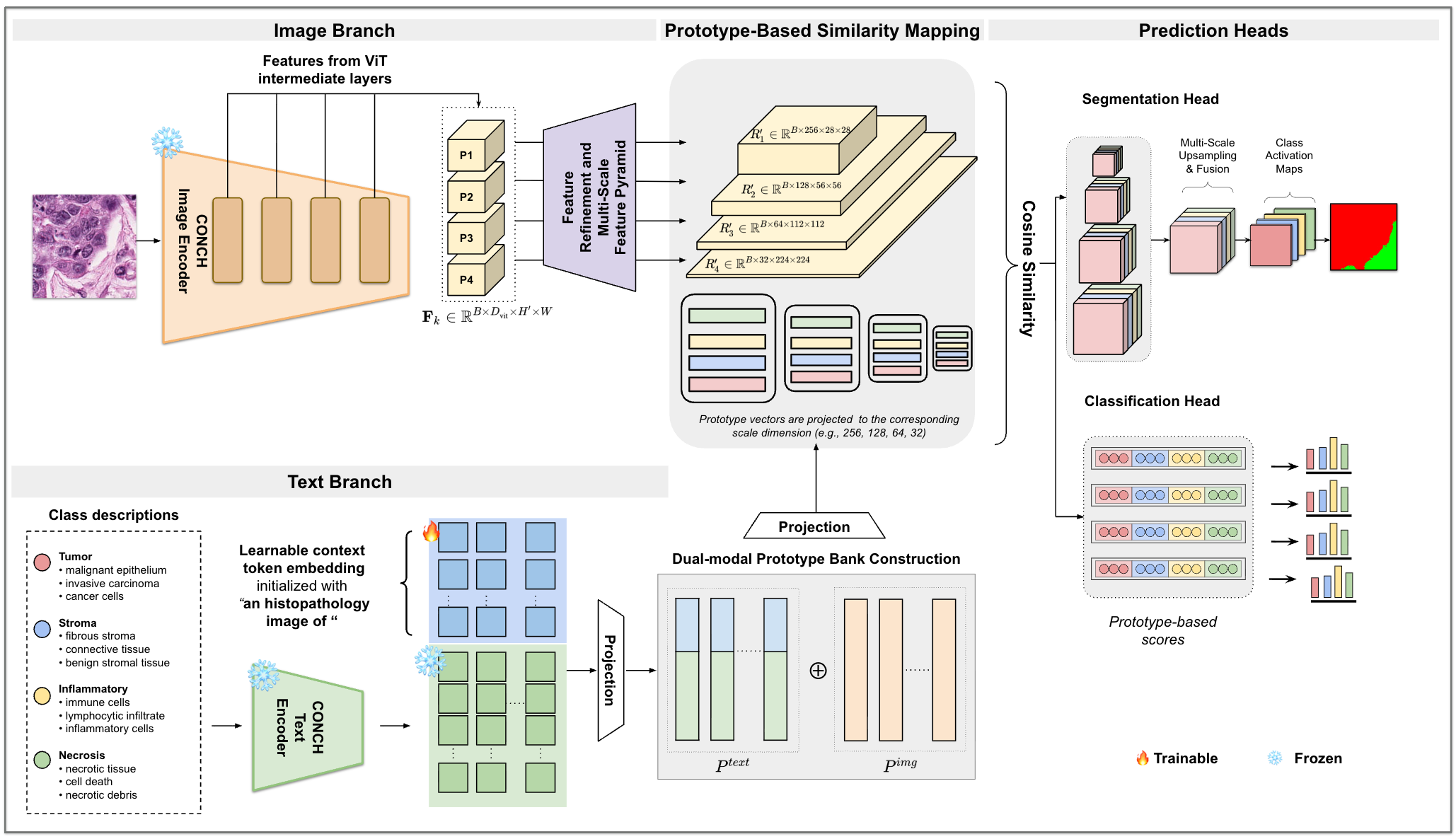}
    \caption{Overview of the Dual-ProtoSeg framework. Multi-scale image features (Image Branch) and prompt-guided text embeddings (Text Branch) form a dual-modal prototype bank, whose similarity to visual features generates multi-scale CAMs for weakly supervised segmentation (Prediction Heads).}     
    \label{fig:overview}
\end{figure}
\subsection{Multi-Scale Visual Feature Extraction and Refinement}

\textbf{Image Feature Extraction.} We use the frozen CONCH ViT-B/16 encoder, pretrained on large-scale pathology image--text pairs, to extract multi-scale visual representations. The transformer naturally produces a hierarchy of hidden states with increasing semantic richness across depth. To capture this information, we extract intermediate hidden states from several representative layers (e.g., 2, 5, 8, 11) using forward hooks. Each feature map is reshaped into a 2D spatial grid and denoted as
\[
\mathbf{F}_k \in \mathbb{R}^{B \times D_{\text{vit}} \times H' \times W'}, \quad k=1,\dots,4,
\]
where $D_{\text{vit}}$ is the token embedding dimension and $H'\!\times\!W'$ corresponds to the patch-token grid.

\textbf{Feature Refinement.} Although semantically meaningful, ViT intermediate features may exhibit blurred boundaries and weak localization. We therefore apply a lightweight refinement module to each $\mathbf{F}_k$. The features are first projected to $D_{\text{ref}}/2$ channels via a $1{\times}1$ convolution with GroupNorm and SiLU, followed by a series of residual blocks with $3{\times}3$ convolutions that improve spatial coherence and suppress noise. A final $1{\times}1$ convolution expands the representation to $D_{\text{ref}}$ channels and prepares it for prototype matching:
\[
\mathbf{R}_k \in \mathbb{R}^{B \times D_{\text{ref}} \times H' \times W'}, \quad k=1,\dots,4.
\]

\textbf{Multi-Scale Feature Pyramid. (Multi-ScaleFP)} To integrate information at different spatial resolutions, the refined maps $\{\mathbf{R}_k\}$ are transformed into a pyramid of features with progressively increasing resolution and decreasing channel dimension. Each $\mathbf{R}_k$ is projected through a lightweight convolutional block and then resized by bilinear interpolation such that spatial dimensions double and channels halve at each level. Formally,
\[
[R'_i]_{i=1}^{4} = \text{Multi-ScaleFP}([\mathbf{R}_i]_{i=1}^{4}),\quad
R'_i \in \mathbb{R}^{B \times D_i \times H_i \times W_i},\;
H_i = 2^{i}H',\;
W_i = 2^{i}W',\;
D_i = D_{\text{ref}}/2^{\,i-1}.
\]

For example, if $\mathbf{R}_i$ has shape $[B,512,14,14]$, the pyramid yields: \(
R'_1: [B,256,28,28],\quad
R'_2: [B,128,56,56],\quad
R'_3: [B,64,112,112],\quad
R'_4: [B,32,224,224].
\) This multi-scale construction produces spatially detailed and semantically consistent feature maps suitable for downstream dense prediction tasks.

\subsection{Dual-Modal Prototype Bank Construction}
\textbf{Text-Based Prototypes with Learnable Context Tokens.} To encode class-specific textual information, we adopt \textit{Learnable Prompt Tokens} called \text{ConchPromptLearner}, inspired by CoOp \cite{zhou2022conditional}, which allows adaptation of class descriptions while keeping the CONCH text encoder frozen. For $C$ classes, each class $c$ is assigned $n_{\text{ctx}}$ learnable context tokens: 
\[
\mathbf{T}_c \in \mathbb{R}^{n_{\text{ctx}} \times D_{\text{text}}},
\] 
where $D_{\text{text}}$ is the token embedding dimension of the CONCH text encoder. Each token sequence $\mathbf{T}_c$ is initialized from a base textual template (e.g., ``a histopathology image of'') and optionally perturbed with Gaussian noise:
\(
\mathbf{T}_c^{(0)} = \mathbf{E}(\text{ctx\_init}) + \boldsymbol{\epsilon}_c, \quad \boldsymbol{\epsilon}_c \sim \mathcal{N}(0, \sigma^2),
\) where $\mathbf{E}$ is the frozen token embedding. The tokens for all classes are stacked into a trainable parameter:
\(\mathbf{T}_{\text{ctx}} = \text{stack}(\mathbf{T}_1, \dots, \mathbf{T}_C) \in \mathbb{R}^{C \times n_{\text{ctx}} \times D_{\text{text}}}.
\)

Each class $c$ has $n_{\text{desc}}$ textual descriptions $\{t_c^{(d)}\}$.  
For each description, the learnable context tokens $\mathbf{T}_c$ are inserted into the prompt (e.g.,  
$[\text{BOS}, \mathbf{T}_c, t_c^{(d)}, \text{EOS}]$), which is padded or truncated to 77 tokens.  
The frozen text encoder processes the prompt, and the end-of-text (EOT) embedding is extracted, linearly projected, and L2-normalized to obtain a text prototype:
\[
\mathbf{f}_{c,d}^{\text{text}}
= \text{normalize}\big(\text{LN}(\mathbf{H}_{\text{EOT}}) W_{\text{proj}}\big).
\]
 
% Each class $c$ can have $n_{\text{desc}}$ textual descriptions, denoted as $\{t_{c}^{(d)}\}_{d=1}^{n_{\text{desc}}}$. For each description, learnable tokens are inserted at a configurable position within the prompt (front, middle, or end), for example at the end:
% \(\tilde{\mathbf{p}}_c^{(d)} = [\text{BOS}, \mathbf{T}_c, t_{c}^{(d)}, \text{EOS}],
% \) with sequences padded or truncated to 77 tokens (CLIP-based setting).   
% During forward propagation, the embedding table is updated with the current token values. Token embeddings are summed with positional embeddings and processed by the frozen text transformer. The end-of-text (EOT) embedding is extracted and projected:
% \[
% \mathbf{f}_{c,d}^{\text{text}} = \text{LN}(\mathbf{H}_{\text{EOT}}) W_{\text{proj}},
% \quad \text{then normalized as: } 
% \mathbf{f}_{c,d}^{\text{text}} \leftarrow \frac{\mathbf{f}_{c,d}^{\text{text}}}{\|\mathbf{f}_{c,d}^{\text{text}}\|_2}.
% \]
At this stage, each textual description produces a separate prototype vector, without averaging across descriptions. The collection of all text-based prototypes is
\[
\mathbf{P}^{\text{text}} = \{\mathbf{f}_{c,d}^{\text{text}} \}_{c=1,\dots,C; d=1,\dots,n_{\text{desc}}} \in \mathbb{R}^{C \cdot n_{\text{desc}} \times D_{\text{text}}}.
\]

During training, only the learnable tokens $\mathbf{T}_{\text{ctx}}$ and a lightweight visual projection layer are updated, while the CONCH backbone remains frozen. The ConchPromptLearner thus constructs a structured Text-based Prototype Bank, with each description forming an individual prototype aligned with visual features.

 \textbf{Learnable Image-based Prototypes.} Let $\mathbf{P}^{\text{img}} \in \mathbb{R}^{C \times D_{\text{img}}}$ denote the set of learnable image prototypes, where $D_{\text{img}}$ is the visual feature dimension (e.g., 512). These prototypes are initialized randomly and optimized during training to complement the text prototypes. Learnable image prototypes are also normalized:
\[
\mathbf{P}^{\text{img}}_{\text{norm}} = \text{normalize}(\mathbf{P}^{\text{img}}) \in \mathbb{R}^{C \times D_{\text{img}}}.
\]

\textbf{Dual-modal Prototype Bank Construction.}
Text and image prototypes are then interleaved per class to form a combined prototype bank:
\[
\mathbf{P}^{\text{combined}} = [\mathbf{P}^{\text{text}}_{\text{proj}, 1}, \mathbf{P}^{\text{img}}_{\text{norm}, 1}, \dots, \mathbf{P}^{\text{text}}_{\text{proj}, C}, \mathbf{P}^{\text{img}}_{\text{norm}, C}] \in \mathbb{R}^{2C \times D_{\text{img}}}.
\]

Finally, to match the multi-scale visual feature dimensions, the combined prototypes are projected through a series of learnable linear layers:
\(
\mathbf{P}^{(s)} = \text{normalize}\big(\mathbf{P}^{\text{combined}} W^{(s)}_{\text{proto}}\big), \quad s = 1, \dots, S,
\)
where $S$ is the number of feature scales (e.g., $S=4$) and $W^{(s)}_{\text{proto}}$ projects to the corresponding scale dimension (e.g., 256, 128, 64, 32). This produces a multi-scale dual-modal prototype bank that jointly represents textual and visual information for each class and is used for attention-based or similarity-based reasoning with image features.

\subsection{Multi-Scale Pseudo-Mask and Segmentation Mask Generation} 
Using the multi-scale dual-modal prototype bank and pyramid features $\{R'_i\}_{i=1}^4$, we generate class activation maps (CAMs) by computing the similarity between each spatial feature and the corresponding projected prototypes at each pyramid level. 

For the $i$-th pyramid level, let $R'_i \in \mathbb{R}^{B \times D_i \times H_i \times W_i}$ be the feature map and $\mathbf{P}^{(i)} \in \mathbb{R}^{K \times D_i}$ the associated prototype bank. We compute cosine similarity with a learnable scaling factor $\text{logit\_scale}_i$:
\(
\text{CAM}_i = \text{logit\_scale}_i \cdot \text{normalize}(R'_i) \cdot \text{normalize}(\mathbf{P}^{(i)})^\top \in \mathbb{R}^{B \times K \times H_i \times W_i}.
\)

This is applied independently at all pyramid levels, producing multi-scale CAMs $\{\text{CAM}_i\}_{i=1}^4$. Each CAM is then upsampled to the original image size $(H_{\text{img}}, W_{\text{img}})$:
\[
\widetilde{\text{CAM}}_i = \text{Upsample}(\text{CAM}_i, \text{size}=(H_{\text{img}}, W_{\text{img}})), \quad i=1,\dots,4.
\]

The upsampled CAMs are fused via element-wise averaging to produce the final pseudo-mask:
\[
\text{CAM}_{\text{fused}} = \frac{1}{4} \sum_{i=1}^{4} \widetilde{\text{CAM}}_i \in \mathbb{R}^{B \times K \times H_{\text{img}} \times W_{\text{img}}}.
\]

During training, we compute multi-label classification losses at each pyramid level by comparing pooled CAMs with image-level labels. The overall loss is a weighted sum of the four levels:
\(
\text{cls\_loss} = \lambda_1 \, \text{loss}_1 + \lambda_2 \, \text{loss}_2 + \lambda_3 \, \text{loss}_3 + \lambda_4 \, \text{loss}_4,
\)
which encourages the model to learn discriminative features at multiple resolutions for weakly supervised segmentation.

To generate the final segmentation mask, we apply post-processing methods that refine boundaries. In particular, we adopt the fully-connected DenseCRF approach from~\cite{CRF} at inference, which enhances spatial coherence and sharpness of the pseudo-mask, producing a high-quality segmentation output.

\section{Experments and Results}
% \textbf{Experiment Settings.} We evaluate our method on the BCSS-WSSS dataset, which contains four classes: TUM, STR, LYM, and NEC. Our model uses the CONCH vision-language backbone. Training is performed with the AdamW optimizer (learning rate $1\times10^{-5}$, weight decay 0.003), a batch size of 64, and for 20 epochs. The checkpoint with the best validation performance is used for evaluation, and segmentation quality is quantified using mean Intersection-over-Union (mIoU) and Dice scores.

\textbf{Experiment Settings.} We evaluate on the BCSS-WSSS dataset with four classes (TUM, STR, LYM, NEC) using a CONCH backbone. Training uses AdamW (lr $1\times10^{-5}$, weight decay 0.001), batch size 64, and 20 epochs. The best validation checkpoint is used for testing, and performance is measured with mIoU and Dice scores.

\textbf{Quantitative Results}. On the BCSS-WSSS dataset (Table~\ref{table:compare}), our method achieves 71.35\% mIoU and 83.14\% mDice, outperforming the previous state-of-the-art (PBIP) by +1.93\% and +1.30\%, respectively (Table~\ref{tab:bcss_wsss_compact_reformat}). It sets new best results for tumor (81.34\% IoU), stroma (68.84\% IoU), and necrosis (69.83\% IoU), with improvements of up to +4.16\% on these clinically critical classes, while achieving the highest lymphocyte Dice (79.08\%). These consistent gains under extremely sparse supervision establish a new state-of-the-art on BCSS-WSSS and enable substantially more accurate delineation of key pathological structures in breast cancer histopathology. 

\begin{figure}[h!]
    \centering
    \includegraphics[width=0.7\linewidth]{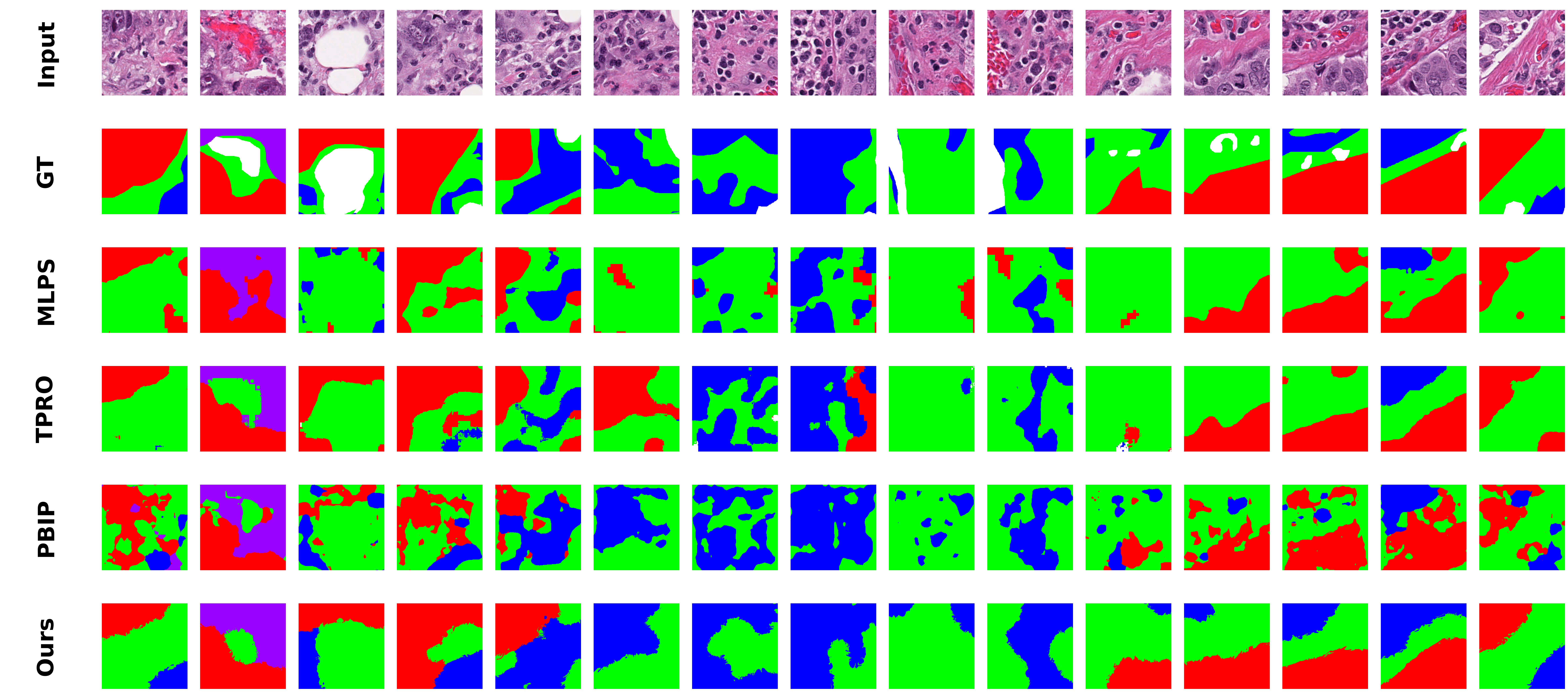} % path + file
    \caption{Qualitative results on BCSS-WSSS test patches. GT denotes ground truth segmentation} 
    \label{fig:qualitative}
\end{figure}

\begin{table*}[h]
\centering
\renewcommand{\arraystretch}{1.15}
\setlength{\tabcolsep}{6pt}
\caption{Segmentation results on \textbf{BCSS-WSSS}}
\label{tab:bcss_wsss_compact_reformat}
\begin{tabular}{@{} l | r r | r r r r | r r r r @{}}
\toprule
\textbf{Method}
& \multicolumn{2}{c|}{\textbf{Metrics (\%)}}
& \multicolumn{4}{c|}{\textbf{Per-class IoU (\%)}}
& \multicolumn{4}{c}{\textbf{Per-class Dice (\%)}} \\
& \textbf{mIoU} & \textbf{mDice}
& \textbf{TUM} & \textbf{STR} & \textbf{LYM} & \textbf{NEC}
& \textbf{TUM} & \textbf{STR} & \textbf{LYM} & \textbf{NEC} \\
\midrule
\textbf{TPRO}
& 65.54 & 78.93
& 77.29 & 66.83 & 56.81 & 61.23
& 87.19 & 80.12 & 72.46 & 75.95 \\
\textbf{MLPS}
& 61.58 & 75.95
& 72.98 & 62.58 & 52.03 & 58.73
& 84.38 & 76.99 & 68.45 & 74.00 \\
\textbf{Proto2Seg}
& 57.42 & 72.24
& 63.25 & 58.28 & 53.27 & 54.89
& 77.49 & 73.64 & 67.78 & 70.08 \\
\textbf{PBIP}
& 69.42 & 81.84
& 77.92 & 64.68 & 65.40 & 69.69
& 87.59 & 78.56 & 79.08 & 82.14 \\
\bottomrule
\textbf{OURS}
& \textbf{71.35} & \textbf{83.14}
& \textbf{81.34} & \textbf{68.84} & 65.39 & \textbf{69.83}
& \textbf{89.71} & \textbf{81.54} & \textbf{79.08} & \textbf{82.23} \\
\bottomrule
\end{tabular}\label{table:compare}
\end{table*} 
\textbf{Qualitative Results.} Figure~\ref{fig:qualitative} presents qualitative segmentation results on representative BCSS-WSSS patches. While all methods capture the general tissue layout, our approach produces noticeably sharper boundaries and clearer structural details that better match the ground truth. Overall, our method delivers more accurate and consistent segmentation of tumor, stroma, necrosis, and lymphocyte regions, reflecting the quantitative improvements reported in Table~\ref{tab:bcss_wsss_compact_reformat}. These results highlight the ability of our approach to generate precise and reliable segmentations under limited weak supervision.

\section{Ablation Studies and Additional Analysis}
\subsection{Complementarity of Text- and Image-based Prototypes.}

Figure~\ref{fig:heatmap} shows that text prototypes activate on broad semantic regions, while image prototypes capture finer visual details recovering missed stroma areas (a, blue) and detecting subtle necrotic fragments in tumor (b, purple). Table~\ref{tab:ablation_img} confirms this: adding image prototypes improves performance from 71.13\% to 71.35\% mIoU and yields the largest gains for lymphocytes. Image prototypes therefore provide complementary fine-grained cues that enhance segmentation beyond what text prototypes alone can offer. 

 \begin{figure}[h]
    \centering
    \includegraphics[width=0.6\linewidth]{figures/heatmap.png} % path + file
    \caption{Complementary activation patterns of text-based (first row) and image-based (second row) prototypes. The first column shows the predicted mask (top) and the ground-truth mask (bottom). (a) Stroma: image prototypes recover missed regions (blue boxes). (b) Tumor: image prototypes detect fine-grained structures (purple boxes).} 
    \label{fig:heatmap}
\end{figure} 

\begin{table*}[h]
\centering
\renewcommand{\arraystretch}{1.18}
\setlength{\tabcolsep}{7.8pt}
\caption{Ablation of image prototypes in the hybrid prototype bank on \textbf{BCSS-WSSS} (10 text descriptions per class). “Yes’’ uses 10 additional image prototypes per class; “No’’ uses only the 10 text prototypes.} 
\label{tab:ablation_img}
\begin{tabular}{@{} c | r r | r r r r | r r r r @{}}
\toprule
\textbf{Image} & \multicolumn{2}{c|}{\textbf{Metrics (\%)}} 
& \multicolumn{4}{c|}{\textbf{Per-class IoU (\%)}} 
& \multicolumn{4}{c}{\textbf{Per-class Dice (\%)}} \\
\textbf{Proto.} & \textbf{mIoU} & \textbf{mDice}
& \textbf{TUM} & \textbf{STR} & \textbf{LYM} & \textbf{NEC}
& \textbf{TUM} & \textbf{STR} & \textbf{LYM} & \textbf{NEC} \\
\midrule
Yes & \textbf{71.35} & \textbf{83.14} & \textbf{81.34} & \textbf{68.84} & \textbf{65.39} & 69.83 & \textbf{89.71} & \textbf{81.54} & \textbf{79.08} & 82.23 \\
No  & 71.13 & 82.98 & 81.14 & 68.50 & 64.37 & 70.52 & 89.59 & 81.30 & 78.32 & 82.71 \\
\bottomrule
\end{tabular}
\end{table*}

\subsection{Influence of Textual Descriptions and Context Length.}
Table~\ref{tab:ablation_num_desc} shows that adding more textual descriptions per class steadily improves performance, as greater linguistic diversity strengthens text–image alignment. Using 10 descriptions achieves the best results (71.35\% mIoU, 83.14\% mDice), with notable gains for stroma and lymphocytes. Table~\ref{tab:ablation_ctx_desc} further shows that context length has minimal effect when only one description is used but becomes impactful with 10 descriptions, where longer contexts consistently improve accuracy and ctx=16 performs best. Overall, description diversity and prompt capacity contribute complementary benefits to segmentation quality.

\begin{table*}[h]
\centering
\renewcommand{\arraystretch}{1.18}
\setlength{\tabcolsep}{7.5pt}
\caption{Ablation study on the number of textual descriptions per class (\#Desc) on \textbf{BCSS-WSSS}}
\label{tab:ablation_num_desc}
\begin{tabular}{@{} c | r r | r r r r | r r r r @{}}
\toprule
\textbf{\#Desc} 
& \multicolumn{2}{c|}{\textbf{Metrics (\%)}} 
& \multicolumn{4}{c|}{\textbf{Per-class IoU (\%)}} 
& \multicolumn{4}{c}{\textbf{Per-class Dice (\%)}} \\
& \textbf{mIoU} & \textbf{mDice}
& \textbf{TUM} & \textbf{STR} & \textbf{LYM} & \textbf{NEC}
& \textbf{TUM} & \textbf{STR} & \textbf{LYM} & \textbf{NEC} \\
\midrule
1  & 70.90 & 82.82 & \textbf{81.36} & 68.48 & 64.77 & 69.01 & \textbf{89.72} & 81.29 & 78.62 & 81.66 \\
3  & 70.86 & 82.82 & 80.27 & 68.02 & 64.84 & \textbf{70.32} & 89.05 & 80.97 & 78.67 & \textbf{82.57} \\
\textbf{10} 
   & \textbf{71.35} & \textbf{83.14} 
   & 81.34 & \textbf{68.84} & \textbf{65.39} & 69.83 
   & 89.71 & \textbf{81.54} & \textbf{79.08} & 82.23 \\
\bottomrule
\end{tabular}
\end{table*}

\begin{table*}[h]
\centering
\renewcommand{\arraystretch}{1.18}
\setlength{\tabcolsep}{7.8pt}
\caption{Ablation of textual description count and context length (ctx) on \textbf{BCSS-WSSS}.}
\label{tab:ablation_ctx_desc}
\begin{tabular}{@{} c | r r | r r r r | r r r r @{}}
\toprule
\textbf{ctx} & \multicolumn{2}{c|}{\textbf{Metrics (\%)}} 
& \multicolumn{4}{c|}{\textbf{Per-class IoU (\%)}} 
& \multicolumn{4}{c}{\textbf{Per-class Dice (\%)}} \\
& \textbf{mIoU} & \textbf{mDice}
& \textbf{TUM} & \textbf{STR} & \textbf{LYM} & \textbf{NEC}
& \textbf{TUM} & \textbf{STR} & \textbf{LYM} & \textbf{NEC} \\
\midrule
\multicolumn{11}{c}{\textit{1 textual description per class}} \\
\midrule
4  & 70.81 & 82.76 & \textbf{81.28} & \textbf{68.51} & \textbf{64.51} & 68.96 & 89.67 & \textbf{81.31} & 78.43 & 81.63 \\
8  & 70.35 & 82.44 & 80.46 & 68.43 & 63.65 & 68.85 & 89.17 & 81.26 & 77.79 & 81.55 \\
16 & \textbf{70.90} & \textbf{82.82} & 81.36 & 68.48 & 64.77 & \textbf{69.01} & \textbf{89.72} & 81.29 & \textbf{78.62} & \textbf{81.66} \\
\midrule
\multicolumn{11}{c}{\textit{10 textual descriptions per class}} \\
\midrule
4  & 70.55 & 82.58 & 80.62 & 68.26 & 63.40 & \textbf{69.92} & 89.27 & 81.14 & 77.60 & \textbf{82.30} \\
8  & 70.73 & 82.69 & 81.27 & 68.47 & 63.31 & 69.87 & 89.67 & 81.28 & 77.54 & 82.26 \\
16 & \textbf{71.35} & \textbf{83.14} & \textbf{81.34} & \textbf{68.84} & \textbf{65.39} & 69.83 & \textbf{89.71} & \textbf{81.54} & \textbf{79.08} & 82.23 \\
\bottomrule
\end{tabular}
\end{table*}

\subsection{Pre-Training Assessment of Textual Descriptions.} 
 \begin{figure}[h]
    \centering
    \includegraphics[width=0.8\linewidth]{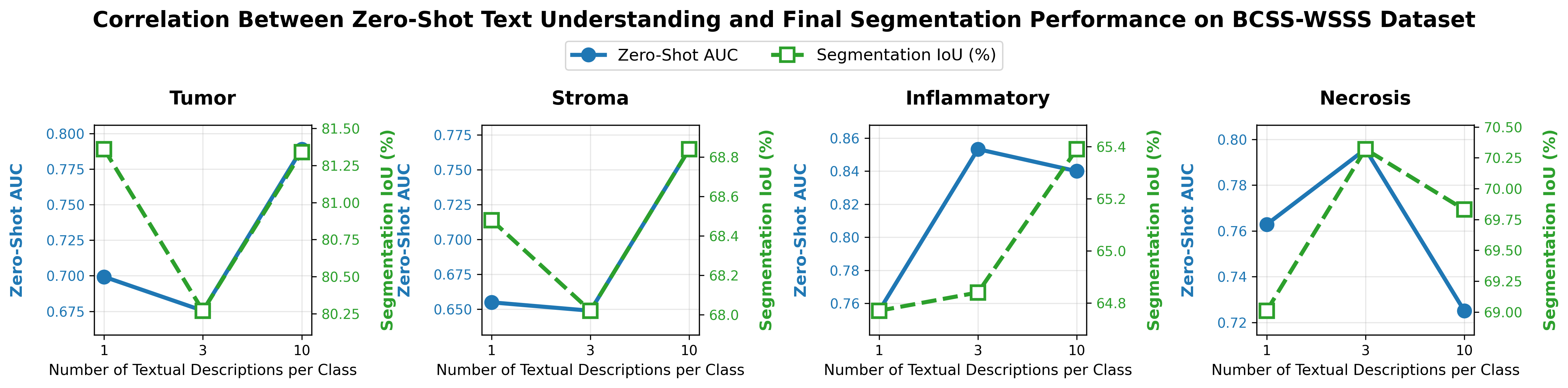} % path + file
    \caption{Relationship between zero-shot text–image alignment (AUC) and downstream segmentation IoU}
    \label{fig:text_understanding_corr}
\end{figure}

Figure~\ref{fig:text_understanding_corr} analyzes textual description quality using zero-shot text–image retrieval AUC from the frozen CLIP text encoder, measuring how well each class’s descriptions align with image features \emph{before} training. Comparing these AUC values with final IoU (using 1, 3, and 10 descriptions) shows that classes with strong initial alignment (tumor, inflammatory) perform well even with few descriptions, while weaker classes (stroma, necrosis) benefit markedly from more diverse descriptions. This suggests that zero-shot AUC is a practical pre-training diagnostic for identifying classes that require richer or more specific prompts.

\section{Conclusion}
We presented a weakly supervised segmentation framework that integrates CLIP-based vision–language features, a dual-prototype design, and a learnable prompt module to improve pseudo-mask quality in histopathology images. The combination of textual and visual prototypes provides complementary cues, enabling more complete coverage of tissue structures, while adaptive textual prompts strengthen alignment between descriptions and image features. Experiments on BCSS-WSSS show consistent improvements across classes and confirm the benefits of richer textual descriptions, longer prompt context, and the inclusion of image-based prototypes. These results demonstrate the effectiveness of multimodal supervision for enhancing weakly supervised segmentation.

% Acknowledgments---Will not appear in anonymized version
\midlacknowledgments{This work was supported in part by the National Institutes of Health (NIH) under Grant 5R01DK134055-02.  }
\label{sec:acknowledgments} 

\bibliography{midl-samplebibliography}

\end{document}